\documentclass{bmvc2k}
\usepackage{comment}
\usepackage{multirow}
\usepackage{amsmath}
\usepackage{subfigure}

\def\cc{P${_{m}}$SC${_{n}}$\xspace}
\def\ccaa{P${_{1}}$SC${_{1}}$\xspace}
\def\ccab{P${_{1}}$SC${_{2}}$\xspace}
\def\ccba{P${_{2}}$SC${_{1}}$\xspace}
\def\ccbb{P${_{2}}$SC${_{2}}$\xspace}
\def\ccbc{P${_{2}}$SC${_{3}}$\xspace}

\newcommand{\red}[1]{\textcolor{black}{#1}}

\title{Parallel Separable 3D Convolution for Video and Volumetric Data Understanding}

\addauthor{Felix Gonda}{fgonda@g.harvard.edu}{1}
\addauthor{Donglai Wei}{donglai@seas.harvard.edu}{1}
\addauthor{Toufiq Parag}{paragt@seas.harvard.edu}{1}
\addauthor{Hanspeter Pfister}{pfister@g.harvard.edu}{1}

\addinstitution{
Harvard John A. Paulson School of Engineering and Applied Sciences\\
Camabridge MA, USA
}

\runninghead{Gonda, Wei, Parag, Pfister}{Parallel Separable 3D Convolution }

\begin{document}
\maketitle
\begin{abstract}
For video and volumetric data understanding, 3D convolution layers are widely used in deep learning,
however, at the cost of increasing computation and training time.
Recent works seek to replace the 3D convolution layer with convolution blocks, e.g. structured combinations of 2D and 1D convolution layers.
In this paper, we propose a novel convolution block, \textit{Parallel Separable 3D Convolution} (\cc), which applies $m$ parallel streams of ${n}$ 2D and one 1D convolution layers along different dimensions. 
We first mathematically justify the need of parallel streams ($P_m$) to replace a single 3D convolution layer through tensor decomposition. 
Then we jointly replace consecutive 3D convolution layers, common in modern network architectures, with the multiple 2D convolution layers ($C_n$).
Lastly, we empirically show that \cc is applicable to different backbone architectures, such as ResNet, DenseNet, and UNet, for different applications, such as video action recognition, MRI brain segmentation, and electron microscopy segmentation.
In all three applications, we replace the 3D convolution layers in state-of-the-art models with \cc and achieve around 14\% improvement in test performance and 40\% reduction in model size and on average.
\end{abstract}
\section{Introduction}\label{sec:intro}

\begin{figure}[t]
  \centering  \centerline{\includegraphics[width=\textwidth]{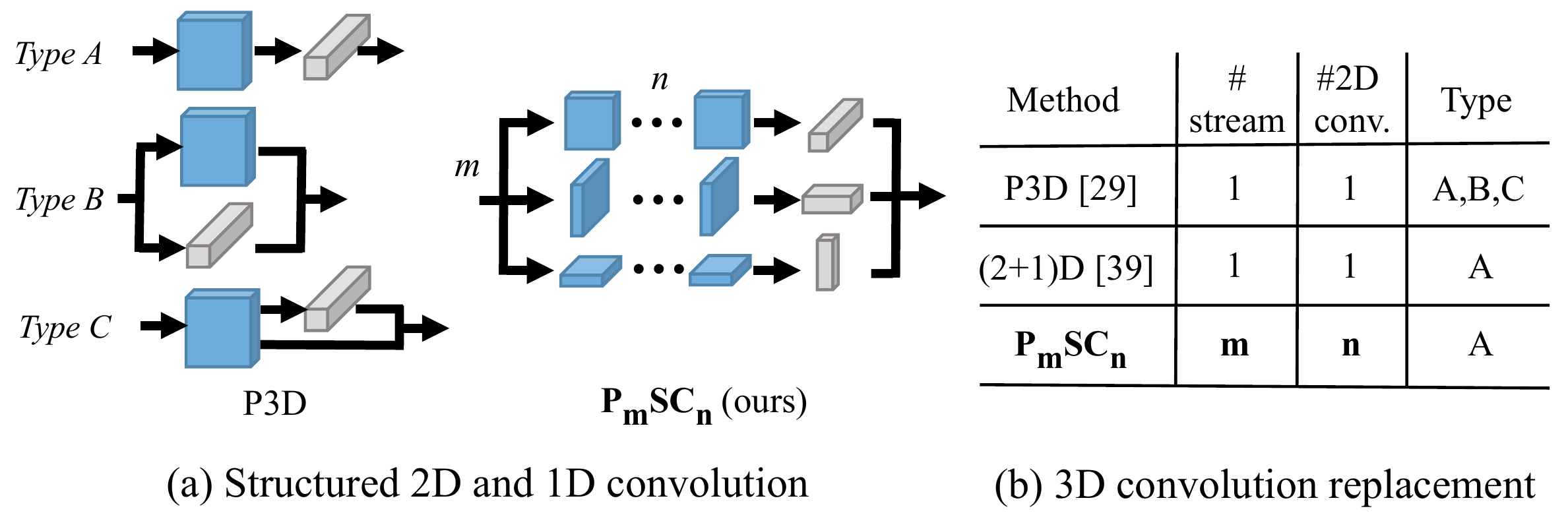}}
\caption{ Illustration of our \textit{parallel separable convolution} (\cc).
The goal is to replace 3D convolution layers with a computational graph of pairs of 2D and 1D convolution layers that are orthogonal to each other.
(a) For layer connection, 
three common types are proposed in Qiu et al.~\cite{qiu2017learning}.
We follow recent works~\cite{tran2017closer} to adopt type A connection and extend it with parallel streams and more 2D layers.
(b) Comparison of our approach with previous work with our categorization.
}
\label{fig:pp3d_conv}
\end{figure}


The advent of deep learning has revolutionized computer vision, especially for problems pertaining to 2D natural images. 
Significant advances have been made to improve 2D convolutional networks, including the design of the convolutional unit~\cite{VeryDeepCNN,He2015Delving,yu15dilated}, the connectivity and scale of the network~\cite{GoingDeeper,huang2016densely, ResNet, srivastava15highway}, and the training strategies~\cite{GlorotAISTATS2010,Kingma14adam,Ioffe15batchnorm}. 

In comparison, the extraction of 3D information from video (time as the third dimension) and volumetric data has just begun to receive increasing attention.
For video understanding, modeling spatial and temporal correlation to capture both the appearance and the dynamics of the video is advantageous for tasks like action recognition. 
For volumetric biological and medical data, 3D contextual information is important for segmenting organs (medical imaging) or cells (biology) collected with different imaging techniques such as computed tomography (CT), magnetic resonance imaging (MRI), and electron microscopy (EM). 

One common approach to capture such 3D context is to use 3D convolution layers.
However, they significantly increase the number of parameters
and complicate the training process.  
Alternatively, recent works in video understanding
propose P3D~\cite{qiu2017learning} and (2+1)D ~\cite{tran2017closer} convolution layers, which  apply 2D spatial and 1D temporal convolution layers in a structured manner with non-linear activations (e.g., ReLU) in between. 
Despite their solid conceptual basis and impressive results on video action recognition, there are three open problems on 3D convolution layers replacement, which we examine in this paper.



First, how does one effectively replace a single 3D convolution layer? 
Previous methods explore the design space of the computation graph of a pair of 2D and 1D convolution layers that operate on orthogonal dimensions. 
Qiu et al.~\cite{qiu2017learning} proposes P3D convolution block with three types of connections (Figure~\ref{fig:pp3d_conv}\red{a}, left) and Tran et al.~\cite{tran2017closer} empirically show that the separable 3D convolutions, i.e. P3D-A, alone can achieve state-of-the-art performance.
We tackle the problem from the tensor decomposition perspective and extend the P3D-A convolution block with $m$ parallel streams using decomposed 2D convolutions in different orientations with mathematical justification.

Second, how does one effectively replace multiple consecutive 3D convolution layers jointly? 
Currently, structured 2D and 1D convolutions is used to replace either each 3D convolution layer independently~\cite{qiu2017learning,tran2017closer}, 
or all 3D convolution layers in the model jointly~\cite{Triplanar,zhou2017fixed}.
We here examine the middle of the spectrum, exploiting the computation redundancy across consecutive layers that are prevalent in modern deep learning architectures, such as VGG-style networks~\cite{simonyan2014very} and ResNet model~\cite{ResNet}. 
We make use of the associative property of convolution operations and propose to add $n$ 2D convolution layers instead of one.

Third, how applicable is such 3D convolution replacement in general?
Recent works~\cite{qiu2017learning,tran2017closer} only apply 3D convolution replacement to ResNet model~\cite{ResNet} for video action recognition. It is unclear how performance varies for different network architectures and applications.
We extensively examine 3D convolution replacement for different architectures such as DenseNet ~\cite{huang2016densely} and UNet~\cite{3DUNET}, and applications, such as MRI brain segmentation and EM neuron segmentation.

In this paper, We propose a generalized separable 3D convolution block, \cc, with multiple successive 2D convolutions along different dimensions followed by 1D convolutions in the complementary dimension (Figure~\ref{fig:pp3d_conv}\red{a}, right).
Regarding the first two problems, we provide theoretical justification for the proposed \cc in Section~\ref{sec:approach}. 
In Section~\ref{sec:app}, to examine the third problem,
we show experimental evidence that \cc outperforms state-of-the-art 3D convolution models for three different architectures in three different applications.
In average, the proposed \cc achieves around 40\% for model size reduction and around 14\% improvement for test performance on average.

\section{Related Work}
\noindent\textbf{Video Action Recognition}
Action recognition is one of the core tasks in video understanding. 
Earlier deep learning works directly apply 2D CNN architectures for image recognition task to different input modalities such as stack of RGB images~\cite{karpathy2014large}, optical flow~\cite{simonyan2014two} and dynamic images~\cite{bilen2017action}.
Much development has been made to improve the feature fusion across different input frames ~\cite{karpathy2014large,ng2015beyond,donahue2015long,wang2016temporal,feichtenhofer2016spatiotemporal,feichtenhofer2017temporal,girdhar2017attentional}. 
Later attempts\cite{c3d} learn motion features end-to-end by 3D convolution filters, but have inferior performance when compared with the two-stream frameworks that encode motion with optical flow. 
More recent work~\cite{feichtenhofer2016convolutional} achieves the state-of-the-art recognition accuracy through a combination of multiple input modalities and 3D convolution.\\

\noindent\textbf{Volumetric Segmentation}
For 3D biological and medical volume segmentation, deep convolutional neural networks have achieved great success. 
In EM connectomics~\cite{morgan13whynot}, where the goal is to discover the biological neural network from enormous volume of EM data, 3D context has played the pivotal role in the recent breakthroughs for neuron segmentation~\cite{lee17superhuman,januszewski16flood}.
In MRI segmentation, Maturana and Scherer~\cite{VoxNet} utilized knowledge of 3D relation to achieve state-of-the-art performance. 
For volumetric cardiac segmentation, Yu~\cite{yu2017automatic} extends
DenseNet by using two dense blocks followed by
pooling layers to reduce feature maps resolution, then restores the resolution by stacks of
learned deconvolution layers.\\

\noindent\textbf{Inference-Time Convolution Approximation}
Given a trained convolution model, inference-time approximation methods aim to compress the learned parameters while achieving similar test performance.
Much work has been done for 2D convolution layer approximation through tensor decomposition. Denton et al.~\cite{denton2014exploiting} use low rank approximation and clustering techniques to approximate a single convolutional layer. 
Mamalet et al.~\cite{mamalet2012simplifying} use rank-1 filters and combine them with an average pooling layer.
Rigamonti et al.~\cite{rigamonti2013learning} show that multiple image filters can be approximated by a shared set of separable filters, which is further explored in Jaderberg et al.~\cite{jaderberg2014speeding} with two schemes of approximation.\\

\noindent\textbf{Training-Time Convolution Replacement}
Given a 3D convolution model design, training-time replacement methods aim to replace each 3D convolution layer with a structured combination of 2D and 1D convolution layers to achieve better test performance after training with similar resource budget. 
For a single 3D convolution layer, Qiu et al.~\cite{qiu2017learning} and Tran et al.~\cite{tran2017closer} use one 2D layer with kernel size ${3 \times 3\times 1}$ in the spatial domain, followed by one 1D layer with kernel size ${1 \times 1\times 3}$ in the temporal domain. They both demonstrate impressive performance improvements on popular action recognition datasets.
For all the 3D convolution layers in the network, Triplanar ConvNet~\cite{Triplanar} utilizes three parallel streams of 2D-version of the original 3D architecture to process orthogonal slices of a 3D volume. The three streams are fused in the final layer to produce a probability map.  In comparison, our \cc convolution can not only approximate each 3D convolution layer, but also a group of consecutive 3D convolution layers that are common in modern architectures, such as ResNet~\cite{ResNet} and DenseNet~\cite{huang2017densely}.

\section{Methods}\label{sec:approach}
In addition to the computation graph explanation (Figure~\ref{fig:pp3d_conv}\red{a}),
we provide mathematical insights of \cc, which can be viewed as the generalized separable 3D convolution.
We first justify the parallel streams through the tensor decomposition of the convolution kernels.
Then we replace consecutive 3D convolution layers with extra number of 2D convolution layers through the commutative and associative property of convolution.
Lastly, we illustrate several parameter choices of \cc and their incorporation into a given neural network model with 3D convolution layers.
\subsection{$m$-Parallel Streams for Single 3D Convolution Layer}\label{subsec:method_parallel}
We sketch the justification for the parallel streams and leave the mathematical details in the supplementary material.
Let us first consider a single convolution kernel $\mathcal{A}$ from the 3D convolution layer, a 4D tensor with size $J_1\times J_2\times J_3\times C$ where $C$ is the number of channels. \\

\noindent\textbf{Separable Convolution Kernel.}
A convolution kernel is called separable if it can be decomposed into the convolution of two or more kernels.
Let $\mathcal{A}^{(k)}$ be a sub-tensor that has size 1 except the $k$-th dimension, e.g. $\mathcal{A}^{(1,2,3)}$ has the size $J_1\times J_2\times J_3\times 1$.
If $\mathcal{A}$ is separable along the first dimension, i.e. decomposable with one 3D sub-tensor $\mathcal{A}^{(2,3,4)}$ and one 1D sub-tensor $\mathcal{A}^{(1)}$,
then for any 4D input tensor $\mathcal{D}$, we have
\begin{align}
\mathcal{A}\ast\mathcal{D}
=(\mathcal{A}^{(1)}\ast\mathcal{A}^{(2,3,4)})\ast\mathcal{D}
=\mathcal{A}^{(1)}\ast(\mathcal{A}^{(2,3,4)}\ast\mathcal{D}), 
\end{align}
\noindent which can be implemented with a chain of 2D and 1D convolution layer.\\
\begin{figure}[t]
  \centering  
  \centerline{\includegraphics[width=\textwidth]{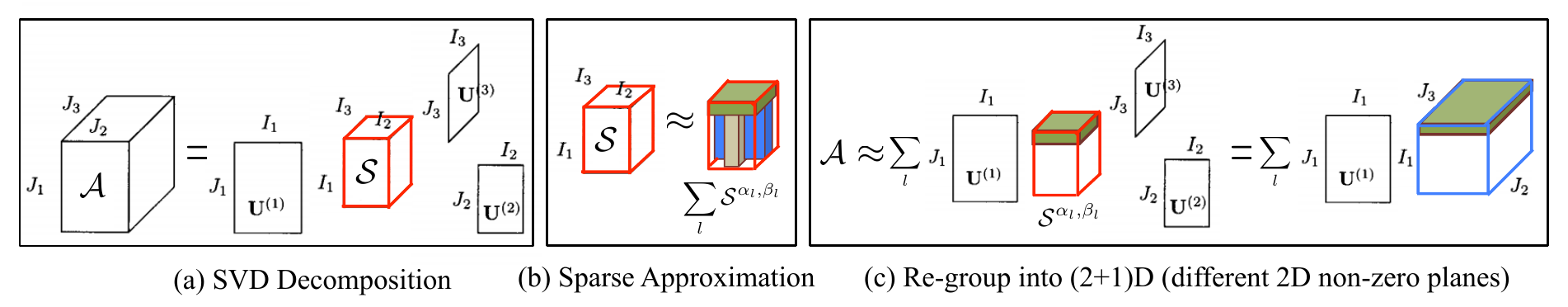}}
\caption{Illustration of the tensor decomposition to justify the need of $m$-parallel streams to replace a single 3D convolution layer. Given one kernel, 4D tensor $\mathcal{A}$, (a) we first decompose it with the singular value tensor $\mathcal{S}$ and orthogonal matrices $\{U^{(k)}\}$ along different dimensions. (b) Then we can rewrite $\mathcal{S}$ as the sum of sub-tensors, (c) each of which leads to a separable convolution kernel along different dimensions. For better model capacity, it is necessary to construct multiple streams of (2+1)D convolution~\cite{tran2017closer}.}\label{fig:decomp}
\end{figure}

\noindent\textbf{General Convolution Kernel.}
Given a general 4D tensor $\mathcal{A}$, 
we show how to decompose it into the sum of separable convolution kernels.
First, we use high-order singular value decomposition (HOSVD)~\cite{lathauwer00multilinear}
to decompose $\mathcal{A}$ with orthogonal matrices $\{U^{(k)}\}_{k\in\{1,2,3\}}$ of the size {\small \{$I_k\times J_k\}_{k\in\{1,2,3\}}$} and singular value tensor $\mathcal{S}$ of the size {\small $I_1\times I_2\times I_3\times C$} (Figure~\ref{fig:decomp}\red{a}).
Then, we further decompose the singular value tensor $\mathcal{S}$ into the sum of tensors $\{\mathcal{S}^{\alpha_l,\beta_l}\}_{\alpha_l\in\{1,2,3\},\beta_l\in\{1,..,I_{\alpha_l}\},}$ whose non-zero entries are only on the $\beta_l$-th sub-tensor along the $\alpha_l$-th dimension.
For example, if $C=1$, then $S$ is a 3D tensor and non-zero elements in each $S^{\alpha_l,\beta_l}$ are on one 2D plane (Figure~\ref{fig:decomp}\red{b}).
\begin{equation}\label{E:s_decomp}
\mathcal{S}=\sum_{l}\mathcal{S}^{\alpha_l,\beta_l}
, ~\mbox{where}~ 
\mathcal{S}^{\alpha_l,\beta_l}_{i_1,i_2,i_3,c}=0~~~\mbox{for }~i_{\alpha_l}\neq\beta_l
\end{equation}
In the supplementary material, we show that each $\mathcal{S}^{\alpha_l,\beta_l}$ leads to a separable convolution kernel $\mathcal{A}^{l}$ (Figure~\ref{fig:decomp}\red{c}), and thus
\begin{align}\label{E:final}
\mathcal{A} =\sum_{l}\mathcal{A}^{l}
=(\sum_{l\in l_1}
\mathcal{A}^{l,(1)}\ast\mathcal{A}^{l,(2,3,4)})
+(\sum_{l\in l_2}
\mathcal{A}^{l,(2)}\ast\mathcal{A}^{l,(1,3,4)})
+(\sum_{l\in l_3}
\mathcal{A}^{l,(3)}\ast\mathcal{A}^{l,(1,2,4)}), 
\end{align}
where $l_k=\{l:\alpha_l=k\}$ grouping $\mathcal{A}^{l}$ by the dimension of their decomposition
\noindent which can be implemented as the sum of three parallel streams of 2D and 1D convolution layers with different orientations.

We claim that the original 3D convolution has much redundancy in model capacity and is vulnerable to overfit training data  (later empirically verified in Figure~\ref{fig:action_experiments}\red{a}). 
By constraining the model to learn separable filters along different dimensions,
\cc not only alleviates the overfitting problem, but also learns 3D context encoded by multi-oriented 2D projections.\\

\noindent\textbf{Adding Non-linearity}
To add non-linearity to the new convolution block,
we add a ReLU layer between the 2D and 1D convolution layer, simliar to the scheme 2 approximation in Jaderberg~\cite{jaderberg2014speeding}.
Instead of summing up different streams, we concatenate them by the channel dimension to further increase the non-linearity.




\begin{figure}[t]
\begin{minipage}[b]{1\linewidth}
  \centering  \centerline{\includegraphics[width=\textwidth]{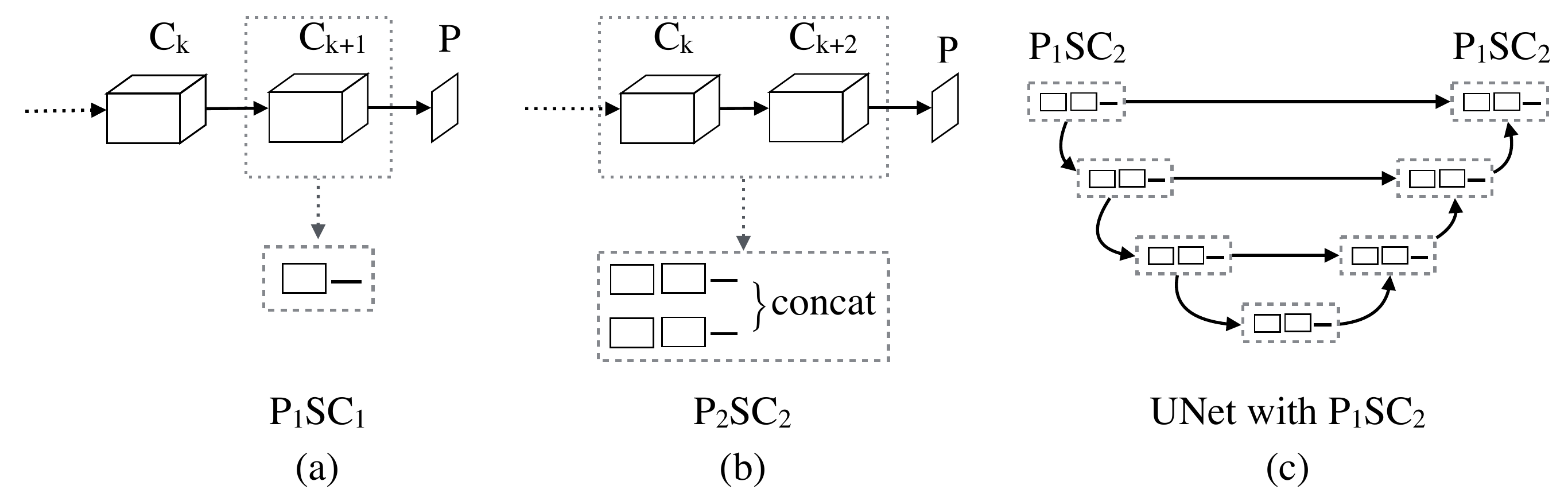}}
\end{minipage}\
\caption{\label{fig:usage} Design examples of how to composite our \cc modules into network segments and whole networks. (a) a \ccaa module replaces a 3D convolution. (b) a \ccbb module replaces a block of two 3D convolutions.  (c) a \ccab module is used to replaced VGG blocks in a U-Net.}
\end{figure}

\subsection{$n$-2D Convolution for Consecutive 3D Convolution Layers}\label{sec:3d-conv-approx}
We provide the intuition on replacing consecutive 3D convolution layers.
Suppose we have two 3D convolution kernel $\mathcal{A}$ and  $\mathcal{B}$ that are separable along the first dimension. Given the commutative and associative property of convolution, we have
\begin{align}\label{E:nconv}
\mathcal{A}\ast\mathcal{B} =
(\mathcal{A}^{(1)}\ast\mathcal{A}^{(2,3,4)})
\ast (\mathcal{B}^{(1)}\ast\mathcal{B}^{(2,3,4)})
=(\mathcal{A}^{(1)}\ast\mathcal{B}^{(1)})\ast \mathcal{A}^{(2,3,4)}
\ast\mathcal{B}^{(2,3,4)}.
\end{align}
Although this is an over-simplified explanation due to the non-linear function in between convolution layers, it suggests a natural extension to include multiple 2D convolution layers instead of one for 3D convolution replacement (Figure~\ref{fig:decomp}\red{a}, right).

\subsection{Parameter Choices}
Our \cc convolution block can be deployed to replace different parts of a neural network model with 3D convolution layers. 
During design, we require the user to choose a value for ${n \in (1,2,...)}$, the dimension of the sub-space, and a value for ${m \in (1,2,3)}$, the number of parallel streams. For ${m > 1}$, the ${n}$-dimension sub-space with a terminal 1D convolution is replicated ${n}$ times and the results of the streams are concatenated as input to the next operation in the network. 
To replace a single 3D convolution layer,
a \ccaa block (Figure\ref{fig:usage}\red{a}) can be used to yield the least amount of changes to a network.  
To replace consecutive 3D convolution layers,
a \ccbb block (Figure\ref{fig:usage}\red{b}) can be used to further reduce computation redundancy. 
Given a neural network model, we limit ourselves to replace convolution layers without pooling layers in between.  
For example, Figure\ref{fig:usage}\red{c} illustrates how to replace the double convolution layers in a UNet architecture with \ccab convolution blocks.

To match the number of parameters of the conventional 3D convolution, we introduce the parameter $M$, to determine the number of filters to use in the sub-space domain. 
\begin{align}\label{eq:m-hyperparam}
	M &= \frac{k_{i}\times k_{i-1} \times d \times d \times d}{(k_{i-1} \times d \times d) + (k_{i} \times d)}
\end{align}
$M$ is adjusted during network construction and is computed using Eqn \ref{eq:m-hyperparam}, where ${k_i}$ is the number of filters of the current layer, ${k_{i-1}}$ is number of filters of the preceding layer, and ${d}$ is the dimension of a symmetric full-rank 3D filter. ${M}$ is scaled by a factor ${\frac{1}{s}}$ so that the number of sub-space filters are evenly distributed across the streams.

\section{Applications}\label{sec:app}
To show general applicability, 
we apply our \cc convolutions to three different applications, 
i.e. action recognition from videos, brain segmentation from MRI images, and neuron segmentation from EM images.
We compare them with the corresponding state-of-the-art baseline models, i.e. ResNet, DenseNet, and UNet.

For each application, our approximations give rise to three architectures using our \ccaa, \ccbb, and \ccbc convolution blocks. In all experiments, we constraint our approximations to ${3\times3\times3}$ convolutions. We train these network architectures from scratch and compare against a baseline architecture and the state of the art.  In all experiments, we employ the Adam~\cite{Kingma14adam} optimizer with learning rate and batch size customized for each application.
The training details are described in the supplementary.

\subsection{3D ResNet: Action Recognition in Videos}
\noindent\textbf{Datasets.}
We conduct this experiment on the popular video action recognition dataset, UCF101~\cite{ucf101}. This data consists of ${13,320}$ videos from 101 action categories and we use the provided split-1 of training and testing. \\

\noindent\textbf{Setup.}
Same as the state-of-the-art (2+1)D convolution model~\cite{tran2017closer}, we adopt the ResNet-34 model~\cite{ResNet} as the backbone architecture.  We replace all 3D convolutions with \cc counterparts and produce three ResNet-34 variants. 
For comparison, we also compare with the previous state-of-the-art method P3D~\cite{qiu2017learning} which is based on ResNet-152.
These networks take as input video clips and predict the class labels of action categories. 
We report the top-1 clip accuracy on the test split as an average over 20 clips to produce the final prediction.
Following the standard practice, we downsample the input clips to ${64\times64}$ in the spatial domain and we sample 64 consecutive frames from each video.\\

\begin{figure}[t]
  \begin{minipage}{0.50\textwidth}    
    \centering
        \includegraphics[width=0.9\textwidth]{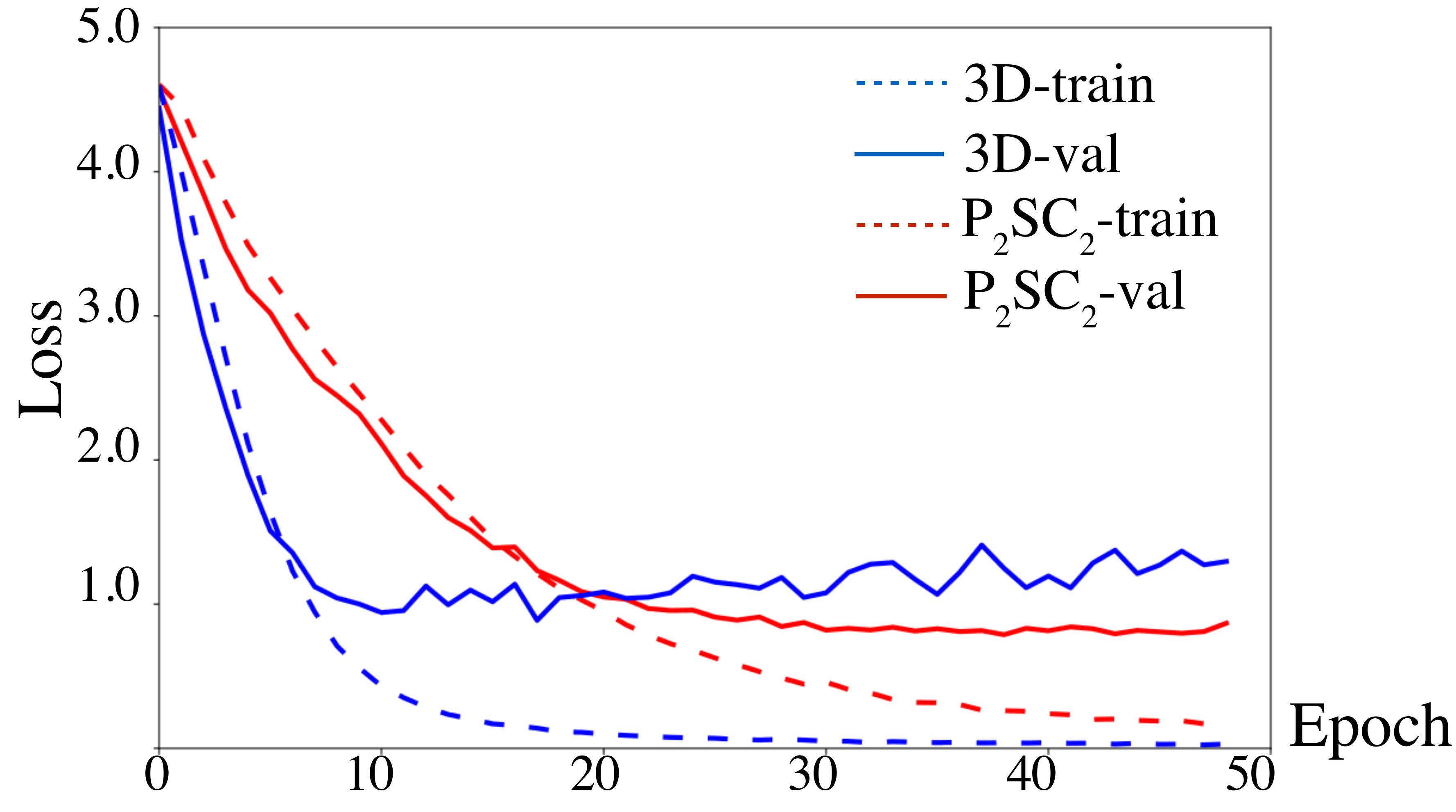}\\
    \vspace{-0.1in}(a)
  \end{minipage}
  \begin{minipage}[c]{0.5\textwidth}
    \centering
        \begin{tabular}{|c|c|c|c|} \hline
        \multicolumn{2}{|c|}{Conv. Type} &  Acc.& \# Param \\\hline        
        \multicolumn{2}{|c|}{P3D$^{\dagger}$~\cite{qiu2017learning}}& 88.6 & 261M\\\hline\hline
        \multicolumn{2}{|c|}{3D}& 85.4 & 64M\\\hline         
        \multirow{3}{*}{\cc} & (1,1)$^{\dagger}$~\cite{tran2017closer} &\textbf{93.6}&39M\\\cline{2-4}
        					 & (1,1) &89.7&39M\\\cline{2-4}
        			         & (2,2) &\textbf{92.3}&49M\\\cline{2-4}
       						 & (2,3) &91.4&33M\\\hline
        \end{tabular}
        \begin{tabular}{c}
        (b)
        \end{tabular}
  \end{minipage}\hfill
  \caption{ 
  Comparison of 3D ResNet models on video action recognition in term of test accuracy and model size.
  We first plot (a) training error (in dashed lines) and validation error (in solid lines) for 3D (blue) and our \cc (red) model for the first 50 epochs.  
  Then we show (b) top-1 accuracy and parameter reduction on 3D ResNet architecture for action recognition.  
  Our \cc variants based on ResNet-34, trained from scratch on UCF-101 dataset, outperform the previous state-of-the-art~\cite{qiu2017learning} (row 1, based on ResNet-152) and original 3D model (row 2). Our best \ccbb model (row 5) is close to Tran et al.~\cite{tran2017closer} (row 3) which is pre-trained on the Sports-1M dataset~\cite{karpathy2014large} (marked with $^{\dagger}$).}
\label{fig:action_experiments}
\end{figure}
\noindent\textbf{Results.}
We first show the training and validation error over epochs for 3D ResNet-34 model with 3D convolution and \ccbb (Figure~\ref{fig:action_experiments}\red{a}).  
The model with 3D convolution quickly overfits the training data and the validation error begins to increase slowly after epoch 10.
In contrast, both the training and validation error decrease steadily for the model with \ccbb, not only reducing model size, but also alleviating the overfitting problem.

For the quantitative comparison, we show the test accuracy of the 3D ResNet-34 model with 3D convolution and \cc variants (Figure~\ref{fig:action_experiments}\red{b}). 
Our \cc variants yield significant reduction in learnable parameters and achieves similar or better accuracy than the baseline architecture.  
However, we note that the published (2+1)D convolutions paper ~\cite{tran2017closer} reports 93.6\% top-1 clip accuracy, which was obtained after pre-training on the Sports-1M dataset~\cite{karpathy2014large}.  
The \cc models are only trained on RGB images from scratch without pre-training. Therefore, despite the different training setting, our results demonstrate that our model can simultaneously achieve better accuracy while remaining efficient. In \ccbb, the number of parameters is larger than the \ccba due to approximating groups of two convolutions in the basic ResNet module.

\subsection{3D DenseNet: Brain Extraction from MRI}
\noindent\textbf{Datasets.}
Our MRI application utilizes T1-weighted MR brain images from the Internet Brain Segmentation Repository (IBSR), which was made available by the Center for Morphometric Analysis, Massachusetts General Hospital\footnotemark. 
The task is to segment brain tissues into four classes of non-overlapping regions:
gray matter (GM), white matter (WM), Cerebrospinal fluid (CSF) and background.
The IBSR dataset is obtained from 18 normal subjects, and the associated manual segmentation is provided by trained experts. We use the manual segmentation as ground truth. 
Following the standard practice, we use eleven subjects for training, five for test, and two for validation.\\
\footnotetext{\url{https://www.nitrc.org}}

\noindent\textbf{Setup.}
3D DenseNet architecture~\cite{densenet} are used to achieve the state-of-the-art results. 
The network architecture consists of five dense blocks, each comprising four convolution layers, followed by a transition layer.  
The original architecture uses a bottleneck residual module with a compression ratio of 0.5 and deconvolution layers with bilinear weight fillers. 
During reproduction, we use the basic 3D residual module without compression and simple up-sampling layers. \\

\begin{figure}[t]
\begin{minipage}{.43\textwidth}
    \subfigure[ ]
    {
        \includegraphics[width=0.67in]{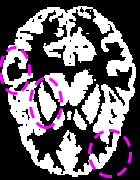}
        \label{fig:mri_gt}
    }\hspace{-0.05in}
    \subfigure[ ]
    {
        \includegraphics[width=0.67in]{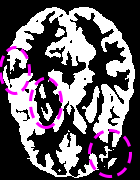}
        \label{fig:mri_3d}
    }\hspace{-0.05in}
    \subfigure[ ]
    {
        \includegraphics[width=0.67in]{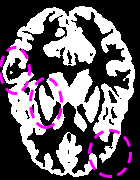}
        \label{fig:mri_psc}
    }
    \label{fig:sample_subfigures}
\end{minipage}\hfill
  \begin{minipage}[c]{0.55\textwidth}
    \centering
        \begin{tabular}{|c|c|c|c|c|c|} \hline
        \multicolumn{2}{|c|}{{\small Conv. Type}} &  {\small WM}& {\small GM} & {\small CSF} & {\small \# Param}  \\\hline
        \multicolumn{2}{|c|}{3D$^{\ddagger}$~\cite{densenet}}& 91.3 & 91.6 & 94.7 & 1.6M\\\hline
        \multicolumn{2}{|c|}{3D}& 85.6 & 88.2 & 84.5 & 5.2M\\\hline
        \multirow{3}{*}{\cc} & (1,1) &95.1  & 94.1 & 93.2&4.7M\\\cline{2-6}
        & (2,2) &\textbf{95.7}  & \textbf{96.1} & 96.3&2.5M\\\cline{2-6}
        & (2,3) &95.2  & \textbf{96.1} & \textbf{97.6}&\textbf{1.4M}\\\hline
        \end{tabular}\\
        \begin{tabular}{c}
        (d)
        \end{tabular}
  \end{minipage}\hfill
  \caption{Comparison of 3D DenseNet models on MRI brain segmentation in term
of test accuracy and model size.
Qualitatively, we show the gray matter (GM) segmentation from (a) ground truth, (b) baseline model, and (c) model with \ccbb (regions within purple eclipse have big difference). Quantitatively, we show the dice coefficients for different regions separately (the higher the better), with white matter (WM) and cerebrospinal fluid (CSF) in addition. 
The proposed \ccbc outperforms previous state-of-the-art model~\cite{densenet} ($\ddagger$: smaller model size due to the usage of bottleneck residual block with compression) in all regions by 4\% in accuracy with 12\% reduction in model size. The 3D model in row 2 uses no compression.
}
  \label{fig:mri_experiments}
\end{figure}

\noindent\textbf{Results.}
For the qualitative comparison,
we show the segmentation of gray matter (GM) from ground truth (Figure~\ref{fig:mri_experiments}\red{a}), DenseNet models using 3D convolution (Figure~\ref{fig:mri_experiments}\red{b}) and \ccbb (Figure~\ref{fig:mri_experiments}\red{c}). 
Moreover, we show purple dotted ellipses to highlight three regions where the proposed method was able to correctly segment the region where the baseline method produces undesirable segments.

For the quantitative comparison, we show Dice scores for the state-of-the-art 3D DenseNet model~\cite{densenet} and our reproduced model with 3D convolution and variants of \cc (Figure~\ref{fig:mri_experiments}\red{d}).
Our best \ccbc model not only consistently achieves around 4\% improvement in segmentation accuracy but also reduces the model size by 12\% compared to the state-of-the-art~\cite{densenet}. The number of parameters in the \ccbb model is less than the \ccaa model due to the DenseNet block, which consists of single convolution.

\subsection{3D UNet: Neuron Segmentation from Electron Microscopy}
\noindent\textbf{Datasets.}
For neuron segmentation, we conducted our experiments on the FIBSEM datasets that were utilized in~\cite{takemura15pnas}.
The FIBSEM datasets are isotropic, i.e., the $x, y, z$ resolutions for each voxel are all same ($10$ nm). The training and test volumes in our experiments have the same size, $500\times 500\times 500$ voxels. \\

\noindent\textbf{Setup.}
3D neuron segmentation usually takes several computational steps.
We here adopts the pipeline in Funke et al.~\cite{funke2017deep}, where a
3D U-Net~\cite{3DUNET} architecture is used in the first step to generate the affinity value for each voxel in x, y, and z direction.
Then watershed and agglomeration methods are used to produce the segmentation from these affinities. 
For a fair comparison, 
we use the same UNet model in the first step and the same set of parameters for later steps as described in Funke et al.~\cite{funke2017deep}.
We train 3D UNet models with original 3D convolution and three variants of \cc until 300k iterations.
We report Variation of Information (VI)~\cite{meila2005comparing}, \cite{parag15iccv} scores to evaluate the segmentation result for each network architecture on the test set.

\begin{figure}
  \begin{minipage}{.55\textwidth}
    \centering
    \subfigure[ ]
    {
        \includegraphics[width=0.82in]{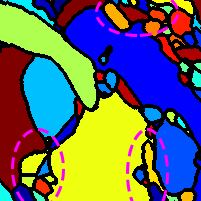}
        \label{fig:em_gt}
    }
    \subfigure[ ]
    {
        \includegraphics[width=0.82in]{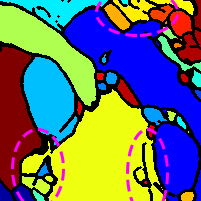}
        \label{fig:em_psc}
    }
    \subfigure[ ]
    {
        \includegraphics[width=0.82in]{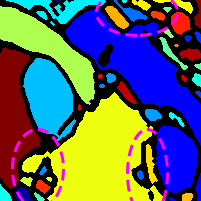}
        \label{fig:em_3d}
    }
    \label{fig:sample_subfigures}
\end{minipage}\hfill
  \begin{minipage}[c]{0.45\textwidth}
    \centering
        \begin{tabular}{|c|c|c|c|} \hline
        \multicolumn{2}{|c|}{{\small Conv. Type}} &  {\small VI}& {\small \# Param} \\\hline        
        \multicolumn{2}{|c|}{{\small 3D~\cite{funke2017deep}}}& (0.10,0.48)& 21M\\\hline         
        \multirow{3}{*}{{\small\cc}} & (1,1) & (0.11,0.24) &11M\\\cline{2-4}
        			    	 & (2,2) & \textbf{(0.07,0.23)} &12M\\\cline{2-4}
       				   {} & (2,3) & (0.08,0.28) &10M\\\hline
        \end{tabular}\\
        \begin{tabular}{c}
        (d)
        \end{tabular}     
  \end{minipage}
\caption{\label{fig:em_results} 
Comparison of 3D UNet models on electron microscopy (EM) neuron segmentation in term of test accuracy and model size.
Qualitatively, we show the dense segmentation result from (a) ground truth, (b) baseline 3D UNet model, and (c) 3D UNet model replaced with \ccbb (regions within purple eclipse have big difference). 
Quantitatively, we show the variational information (VI) scores (the lower the better).
The proposed \ccbb outperforms previous state-of-the-art model~\cite{funke2017deep} by 40\% in accuracy with 40\% reduction in model size.
}
\end{figure}

\noindent\textbf{Results.}
For the qualitative comparison,
we show the segmentation results from ground truth (Figure~\ref{fig:em_results}\red{a}), UNet models using 3D convolution (Figure~\ref{fig:em_results}\red{b}) and \ccbb (Figure~\ref{fig:em_results}\red{c}). 
Moreover, we show purple dotted ellipses to highlight three regions where the proposed method was able to correctly segment the neuron cell region where Funke et al.~\cite{funke2017deep} falsely merge small segments.

For the quantitative comparison,  we show both the under and over-segmentation VI values respectively in parenthesis in the table (Figure~\ref{fig:em_results}\red{d}). Our best \cc approximation with $(m,n)=(2,2)$ reduces both over and under-segmentation and cuts the over-segmentation error by half with respect to the state of the art~\cite{funke2017deep} with 40\% reduction in model size.  
The \ccbb model has more parameters than \ccba due to approximation of groups of two convolutions in the VGG module of the U-Net architecture. 
All our \cc variants reduce the false split error in a similar range without any practical increase in false merge error. 

\section{Conclusions}
We have demonstrated that redundancies in 3D convolution operations can be exploited by using parallel streams of separable convolution filters of 2D and 1D convolutions. 
We presented different ways to combine our \cc modules to approximate convolutions for each single layer or consecutive layers. 
The resulting approximations are computationally efficient while achieving better accuracy on test data. 
Our method is flexible and can be applied to optimizing networks with minimal changes from state-of-the-art deep learning models. 
We make our tensorflow code available at: \href{http://www.rhoana.org/psc}{www.rhoana.org/psc}

\section*{Acknowledgemets}
This work is partially supported by NSF grants IIS-1447344 and IIS-1607800 and the Intelligence Advanced Research Projects Activity (IARPA) via Department of Interior/Interior Business Center (DoI/IBC) contract number D16PC00002.

\bibliography{refs}
\end{document}